\documentclass[sigplan]{acmart}

\acmConference[]{}
\acmYear{2019}
\startPage{1}

\setcopyright{none}

\usepackage{booktabs}   
\usepackage{subcaption} 
\usepackage{array}
\usepackage{enumitem}
\usepackage{makecell}

\begin{document}

\title{A Case Study: Exploiting Neural Machine\\Translation to Translate CUDA to OpenCL}         


\author{Yonghae Kim}
\affiliation{
  \department{College of Computer Science}              
  \institution{Georgia Institute of Technology}            
  \city{Atlanta}
  \state{GA}
  \country{USA}                    
}
\email{yonghae@gatech.edu}          

\author{Hyesoon Kim}
\affiliation{
  \department{College of Computer Science}              
  \institution{Georgia Institute of Technology}            
  \city{Atlanta}
  \state{GA}
  \country{USA}                    
}
\email{hyesoon@cc.gatech.edu}         

\begin{abstract}
The sequence-to-sequence (seq2seq) model for neural machine translation has significantly improved the accuracy of language translation. There have been new efforts to use this seq2seq model for program language translation or program comparisons. In this work, we present the detailed steps of using  a seq2seq model to translate CUDA programs to OpenCL programs, which both have very similar programming styles. Our work shows (i) a training input set generation method, (ii) pre/post processing, and (iii) a case study using Polybench-gpu-1.0, NVIDIA SDK, and Rodinia benchmarks.

\end{abstract}



\maketitle

\section{Introduction}

With the recent development of neural machine translation (NMT), NMT becomes an attractive option for program language translation~\cite{treetotree, ZuoNMT, Bahdanau2015, Cho2014}. Especially, the language agnostic neural network design in sequence-to-sequence (seq2seq) models~\cite{seq2seq_model} is a promising method, as it is currently used in Google. The seq2Seq model can be trained using only a paired translated input without considering the language grammar difference, which makes the same network applicable for different natural languages. 

Table~\ref{tab:comp} shows a comparison between natural languages and programming languages. There are enough similarities but quite a few differences as well. The biggest difference is that natural language translation tolerates grammar errors but program language translation does not. However, that can be corrected through post-processing similar to program debugging after compilation.

	\begin{table*}[!]
	    \centering
	    \caption{Comparison between natural languages and programming languages.}
	    \label{tab:comp}
	    \begin{tabular}{|c||c|c|}
	    \hline
	         &  Natural Language & Programming Language    \\\hline \hline
            Token & 	Punctuation marks	 & Punctuation marks,
                operators, variables \\  \hline
                Scope in sentence	 & Vague	& Explicit  \\ \hline
                Operator precedence & 	None	 & Exist\\  \hline 
                Ambiguous & 	Allowed	& Not allowed \\ \hline
                Naming Scope &  weak  & Explicit \\ \hline 
                Rigid syntax & 	Exist but flexible  in context	 & Exist  \\  \hline
                Number of names	 & Huge but finite & 	Arbitrary  \\ \hline
	    \end{tabular}
	\end{table*}


In this work, we present a case study of source-to-source translation using neural machine translation (NMT) techniques by translating  CUDA to OpenCL. CUDA and OpenCL are both parallel computing programming languages for accelerators. We choose these languages because they share many similarities and provide a good platform to develop translation techniques using NMT and to understand the limitations. Based on the knowledge and techniques from this translation, future work can be expanded to support other program language translation. 

The summary of our contribution is as follows. We develop a dataset generation flow for translating CUDA to OpenCL using NMT. To do so, we first write a pair of API usages for CUDA and OpenCL. With the API usages written, we construct usage symbol trees to extract sentences from the CUDA samples and find uncovered API usages. Using the dataset we generated, we train the NMT system, which learns the structural similarity between CUDA and OpenCL, and translate CUDA code to OpenCL code by inference. Finally, we present translation examples and discuss the limitations and feasibility of our current approach.

\begin{figure}[h]
\centering
\begin{subfigure}{0.20\textwidth}
\centering
\includegraphics[width=0.8\textwidth]{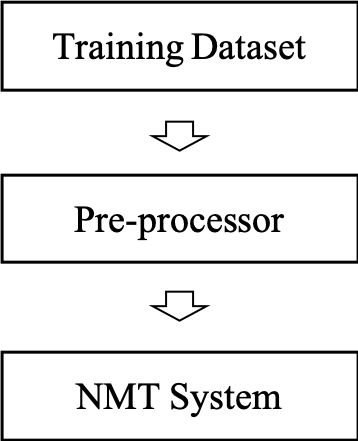}
\end{subfigure}
\begin{subfigure}{0.20\textwidth}
\centering
\includegraphics[width=0.8\textwidth]{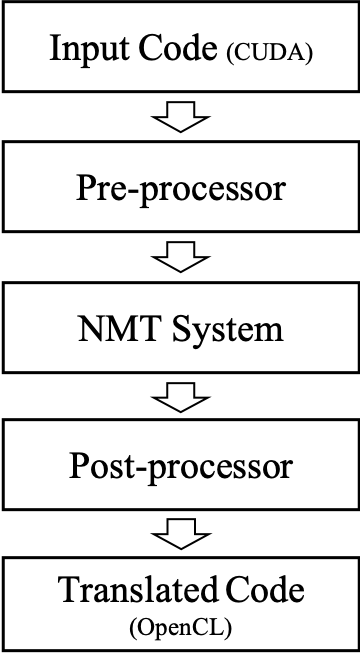}
\label{fig:training}
\end{subfigure}

\begin{subfigure}{0.20\textwidth}
\centering
\caption{training phase}
\end{subfigure}
\begin{subfigure}{0.20\textwidth}
\centering
\caption{inference phase}
\label{fig:inference}
\end{subfigure}
\caption{Overview of workflow \cite{Kim}.}
\label{fig:workflows}
\end{figure}

\section{Background}
\subsection{Neural Machine Translation (NMT)}
Neural networks have demonstrated outstanding performance in natural language processing (NLP). For example, the sequence-to-sequence (seq2seq) model \cite{seq2seq} presents a large, deep Long Short-Term Memory (LSTM), outperforming a mature SMT system by a sizable margin. It also shows the capability of translating very long sentences.

Motivated by the meaningful success and the similarities between natural and programming languages, we exploit the seq2seq model to translate programming languages. We take a statement (or statements) as a sentence and translate it to another sentence written in the target language. Since we translate CUDA to OpenCL, in the training dataset, CUDA code becomes source sentences, and OpenCL code becomes target sentences. During an inference phase, we take as input CUDA code and infer OpenCL code.

\subsection{CUDA vs. OpenCL}
Both CUDA and OpenCL have host code and kernel code. In the host code case, most host API functions have one-to-one correspondence between CUDA and OpenCL. Consider an example of {\tt cudaMalloc} and {\tt clCreateBuffer} that have the same meaning between CUDA and OpenCL. Since a function call of {\tt cudaMalloc} contains all the necessary  information to be translated, such as arguments, we can write a function call of {\tt clCreateBuffer} with a given source code.

Kernel code also has similarities between CUDA and OpenCL. Kernel qualifiers and built-in functions have equivalents, and therefore we replace one program language's keyword with an equivalent one in the other language. Program translation rules between CUDA and OpenCL have already been studied \cite{cu2cl}, \cite{snucu2cl}, and we used these rules to train NMT. 
\section{Overview of Workflow}
In this section, we describe the overview of the NMT workflow. As can be seen in  Fig. \ref{fig:workflows}, in addition to an NMT system, we use a pre-processor during the training phase and a pre-/post-processor during the inference phase. Compared to natural languages in which a large, but finite, set of vocabulary exists, programmers use numerous arbitrary variable names such as alphabet characters or abbreviations of variables or functions. By having a pre-/post-processor, we enable arbitrary variable names in programming languages to be translated. This also contributes to minimizing the size of the vocabulary. Both pre-/post-processors are developed as Python scripts, which are explained as follows.

\subsection{Pre-processor} \label{preprocessor}
A pre-processor performs lexical analyses and variable renaming. First, it reads and tokenizes a given program code. While tokenizing the code, it merges tokens comprise one statement as a sentence---i.e., even when a statement is written in multiple lines, it puts the tokens together as a sentence. In this way, the pre-processor generates a set of sentences, each of which consists of tokens. Next, we map tokens (from tokenized sentences) to three types of symbols depending on their variable type. This is because we try to reduce the cases in which different sentences are renamed to one sentence. As can be seen in Table \ref{table:rename}, we map identifier, string literals, and numeric constants to \textit{\_id}, and each symbol is tagged as a number in ascending order starting from zero---i.e., the first token mapped to \textit{\_id} becomes \textit{\_id0}, and the next one becomes \textit{\_id1}. And, we map operators to \textit{\_op} and data types to \textit{\_tp} with a tagged number in ascending order starting from zero. The numbering for each type has its own order.

Moreover, pre-processing sentences creates a mapping table that contains mapping information between variable names and abstract symbols. This is used later by a post-processor when it replaces the renamed tokens with their original names. Note that we do not rename CUDA/OpenCL APIs since they decide the context of a sentence and which rule to be used to translate it. Finally, the pre-processor is also capable of generating a training dataset. The details of how we generate a dataset are covered in Section \ref{datasetgen}.

	\begin{table}
	    \centering
	    \caption{Mapping tokens to abstract symbols.}
	    \label{table:rename}
	    \begin{tabular}{|c||c|}
	    \hline
	        Symbol & Token    \\\hline \hline
            \_id & identifiers, string literals, numeric constants \\  \hline
            \_op & operators  \\ \hline
            \_tp & data types \\  \hline 
	    \end{tabular}
	\end{table}

\subsection{NMT System}
In an NMT system, we exploit a seq2seq model as it has demonstrated outstanding performance in translating long sentences. The NMT system is a component where actual translation occurs. During a training phase, a training dataset, which consists of source sentences for CUDA and target sentences for OpenCL, is pre-processed and fed into the NMT system. During an inference phase, input CUDA code is pre-processed and fed into the NMT system. Then, the NMT system outputs renamed OpenCL code.

\begin{figure*}[t]
\centering
\includegraphics[width=0.8\textwidth]{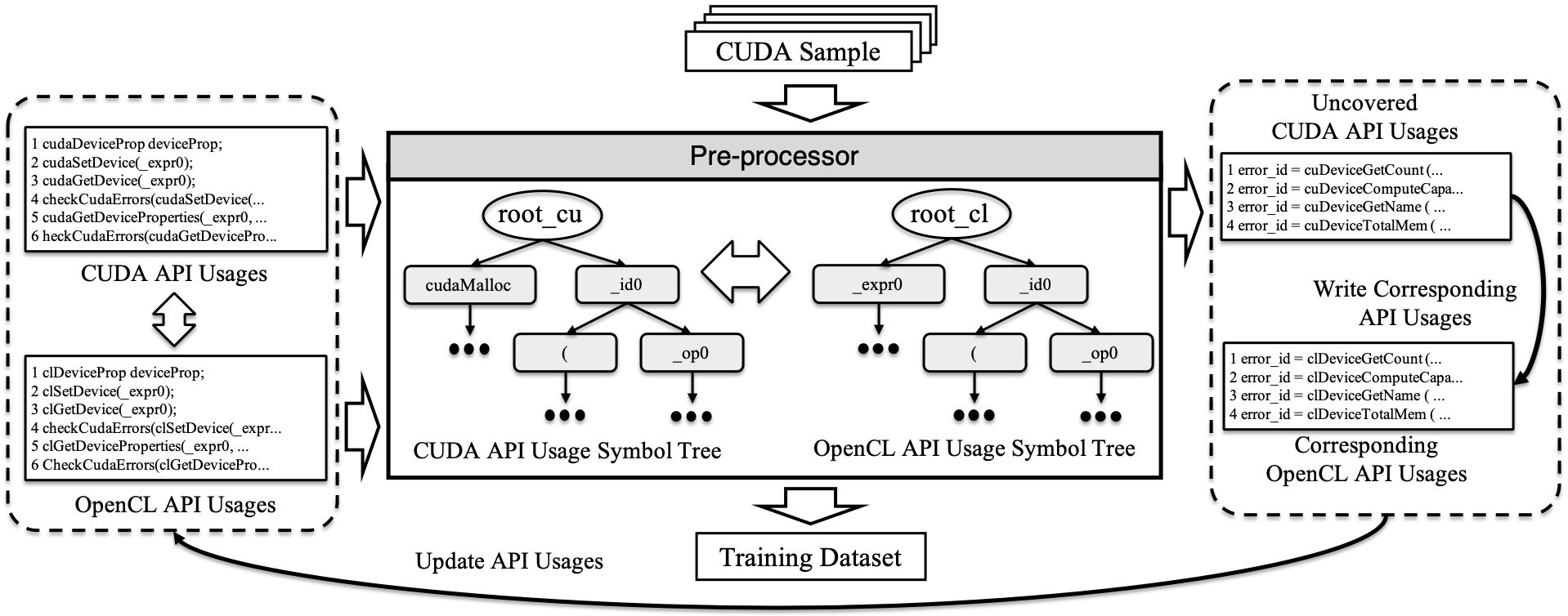}
\caption{Overview of the proposed dataset generation flow.}
\label{fig:gen_flow}
\end{figure*}

\subsection{Post-processor} \label{postprocessor}
A post-processor performs initial name replacement and code restructure. As we train the NMT system with renamed sentences, the output sentences generated from the NMT system also have abstract symbols. Based on the mapping table created by a pre-processor, a post-processor replaces them with their original names. Since the initial variable names from the input code remain in the output code, the translation provides a high-quality code. Finally, based on syntactic rules, it puts appropriate indents between tokens for better readability and generates the final outcome.
\section{Dataset Generation} \label{datasetgen}
Since there is no publicly available dataset for translating CUDA to OpenCL using NMT, we develop a dataset generation flow and generate a dataset from CUDA samples. Fig. \ref{fig:gen_flow} shows the overview of our proposed dataset generation flow, and we explain the steps of dataset generation as below.

\subsection{Steps of Dataset Generation}
We first summarize how we generate a dataset and describe additional details in the following subsections.

\textbf{1) Write API usages}
\begin{itemize} [leftmargin=10pt]
\item Write a pair of API usages for CUDA and OpenCL.
\end{itemize}

\textbf{2) Build usage symbol trees}
\begin{itemize} [leftmargin=10pt]
\item Read API usages written.
\item Tokenize each sentence and rename tokens as abstract symbols.
\item Build usage symbol trees that consist of renamed tokens.
\end{itemize}

\textbf{3) Gather expressions and find uncovered API usages from CUDA samples}
\begin{itemize} [leftmargin=10pt]
\item Tokenize each sentence in CUDA samples.
\item If a sentence includes CUDA APIs, see if the sentence is found in the CUDA usage symbol tree.
\item If found, and the API usage has expression nodes, add each expression to a corresponding expression node.
\item If not found, write the sentence to a separate file that contains uncovered API usages.
\end{itemize}

\textbf{4) Generate a dataset}
\begin{itemize} [leftmargin=10pt]
\item Read again API usages written.
\item If an API usage has expression keywords, permutate the expression nodes and add multiple sentences to a dataset.
\item If not, add sentences to a dataset without permutation.
\end{itemize}

\subsection{API Usage Generation}
Our method requires users to manually write a pair of API usages for CUDA and OpenCL. A CUDA API usage has a sentence pattern to translate, and any variable names can be used in the sentence pattern since they will be renamed as abstract symbols later by a pre-processor. An OpenCL API usage has a sentence pattern that the corresponding CUDA API usage is translated to. Each line of API usages has a one-to-one correspondence with each other---i.e. the first line in the set of CUDA API usages corresponds to the first line in the set of OpenCL API usages.

\subsection{Building a Usage Symbol Tree}
A pre-processor tokenizes API usages manually written and renames tokens as abstract symbols. Then, it builds usage symbol trees that consist of renamed tokens for each CUDA and OpenCL. By traversing the usage symbol trees with tokens in a sentence, we can easily determine whether a given sentence is covered by our API usages, and based on the outcome we can generate a corresponding OpenCL sentence. If a sentence includes CUDA APIs but is not found in the usage symbol tree, it is considered an uncovered API usage and written to a separate file. The usage symbol trees enable us to maintain sentence patterns to translate and easily generate a new larger dataset when we get new CUDA samples.

\subsection{Using an Expression Keyword}
\begin{figure}[t]
\centering
\includegraphics[width=0.2\textwidth]{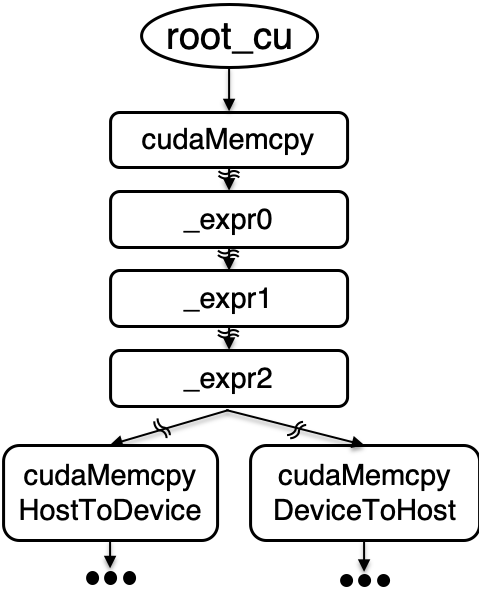}
\caption{Node representation of {\tt cudaMemcpy} usage.}
\label{fig:expr_keyword}
\end{figure}

When we write a pair of API usages, we use an expression keyword, \textit{\_expr}, to represent function parameters. Each one is tagged as a number in ascending order starting from zero. Each expression keyword becomes an expression node in a usage symbol tree. Fig. \ref{fig:expr_keyword} shows the example of node expression using expression keywords. This reduces manual efforts to write API usages. Consider the example of {\tt cudaMemcpy}. It takes four parameters in its function call, and each parameter can have various shapes. Fig \ref{fig:using_expr} (a) shows three different sentences. For these sentences, instead of writing three CUDA API usages for each sentence, we write one CUDA API usage with expression keywords, as shown in Fig \ref{fig:using_expr} (b).

\begin{figure}[t]
\centering
\small
\label{tabular:1}
\begin{tabular}{| m{8cm} |} 
\hline
cudaMemcpy(A\_gpu, A, sizeof(double)*NI, cudaMemcpyHostToDevice); \\
cudaMemcpy(B\_gpu, B, sizeof(double)*NI*NL, cudaMemcpyHostToDevice); \\
cudaMemcpy(C\_gpu, C, sizeof(double)*NI*NJ*NK, cudaMemcpyHostToDevice); \\
\hline
\end{tabular}
(a) Three different {\tt cudaMemcpy} sentences
\label{tabular:2}
\begin{tabular}{| m{8cm} |} 
\hline
cudaMemcpy(\_expr0, \_expr1, \_expr2, cudaMemcpyHostToDevice); \\
\hline
\end{tabular}
(b) {\tt cudaMemcpy} usage with expression keywords

\caption{Example of using expression keywords.}
\label{fig:using_expr}
\end{figure}

\subsection{Expression Node Permutation}
Expression node permutation is used to increase the size of a dataset. As explained in Section \ref{datasetgen}, when we define a pair of API usages, we use an expression keyword. Each expression keyword becomes an expression node in a usage symbol tree. When we generate sentences from CUDA samples, instead of simply finding sentences covered by API usages and adding them to a dataset, we collect expressions to the corresponding expression node. After looking through all samples, we permutate each expression node and generate the increased number of sentences. Consider {\tt cudaMemcpy} in Fig. \ref{fig:expr_keyword} and assume that we collect \textit{n} expressions for \textit{\_expr0}, \textit{m} expressions for \textit{\_expr1}, and \textit{k} expressions for \textit{\_expr2}. By permutating each expression node, we produce $\textit{n} \times \textit{m}$ $\times \textit{k}$ sentences. Note that \textit{\_expr} symbols are not present in sentences fed into the NMT model. It is only used to generate a dataset and is not used as a vocabulary.

	\begin{table}
	    \centering
	    \caption{The number of sentences generated from CUDA benchmarks.}
	    \label{table:number}
	    \begin{tabular}{|c||c|c|c|}
	    \hline
	        Benchmark & \# application & \thead{\# sentences \\ found} & \thead{\# sentences \\ generated} \\    \hline \hline
            Polybench-gpu & 15 & 169 & 221 \\  \hline
            NVIDIA SDK & 25 & 265 & 583  \\ \hline
            Rodinia & 13 & 286 & 538 \\  \hline 
            Total & 53 & 715 & 1874 \\  \hline 
	    \end{tabular}
	\end{table}

	\begin{table}
	    \centering
	    \caption{Hyper-parameters of the NMT system~\cite{seq2seq_model}.}
	    \label{table:hyper}
	    \begin{tabular}{|c||c|}
	    \hline
	        & Seq2Seq model \\    \hline \hline
            Batch size & 128 \\  \hline
            Number of RNN layers & 43  \\ \hline
            RNN cell & LSTM \\  \hline 
            Initial learning rate & 0.005 \\  \hline 
            Dropout rate & 0.2 \\  \hline
            Attention model & scaled luong \\  \hline
	    \end{tabular}
	\end{table}
\section{Results}
As discussed in Section \ref{datasetgen}, we extract sentences from CUDA samples and generate a dataset. Our current dataset has 126 lines of API usages, and we use Polybench-gpu-1.0 \cite{polybench}, NVIDIA SDK \cite{nvidia} examples, and Rodinia \cite{rodinia} benchmarks as target translation. We currently support API usages in the evaluated benchmarks. Table \ref{table:number} presents the number of sentences generated. In the table, Column 2 indicates the number of sentences that include CUDA APIs and are found in our current usage symbol trees, and Column 3 indicates the number of sentences that we generate using the expression node permutation method. With only 126 lines of API usages, we generate 1874 sentences from CUDA samples. We can see that the permutation method increases the number of sentences by 2.6x.

We use the seq2seq model as our NMT model. Table \ref{table:hyper} shows the hyper-parameters of the NMT system used. We use the same dataset as training, development, and test dataset. This is because we intend to have the NMT system to produce a correct sentence and understand the limitation of this NMT-based translation approach. In contrast to natural languages, programming languages have a rigid syntax, and therefore we need to generate correct sentences to make the translated code executable. Therefore, we intentionally cause overfitting by using a shared dataset. While we achieve a 99.1 BLEU score, this does not imply that the NMT system has a better capability to infer and translate unseen sentences. However, it can correctly translate most of the sentences covered by the API usages written by users. For the polybench-gpu-1.0 benchmark, we manually changed about 10 lines out of 200-350 lines for each application and were able to make it run. To fully utilize the potential of using machine learning, we would need a larger dataset and need to split the dataset so that there is no overlap among training, development, and test dataset. We leave this as future work.

Fig \ref{fig:translation_ex} shows a translation example of a 2-D matrix multiplication. Fig. \ref{fig:translation_ex} (a) and (b) show the CUDA host and kernel code, respectively. Before being fed into the NMT system, they are pre-processed first by a pre-processor; Fig. \ref{fig:translation_ex} (c) and (d) show the pre-processed code. If a sentence does not have CUDA APIs, it becomes {\tt\_line\_not\_to\_translate} symbol and is replaced with the original sentences later by a post-processor. After translation using the NMT system, the pre-processed code is translated to OpenCL code that retains the renamed tokens, which can be seen in Fig. \ref{fig:translation_ex} (e) and (f). Finally, a post-processor replaces the renamed tokens with their initial names and provides the final OpenCL code. The final output code is shown in Fig. \ref{fig:translation_ex} (g) and (h).

\begin{figure}
\small
\centering
\begin{tabular}{| m{8cm} |} 
\hline
void mm2Cuda(float* A, float* B, float* C) \{ \\
  float *A\_gpu; \\
  ... \\
  cudaMalloc((void **)\&A\_gpu, sizeof(float) * NI * NK); \\
  cudaMemcpy(A\_gpu, A, sizeof(float) * NI * NK, cudaMemcpyHostToDevice); \\
  ... \\
\hline
\end{tabular}
\label{tabular:2a}
(a) CUDA host code
\begin{tabular}{| m{8cm} |} 
\hline
\_\_global\_\_ void mm2\_kernel1(float *A, float *B, float *C) \{ \\
int j = blockIdx.x * blockDim.x + threadIdx.x; \\
... \\
for (k = 0; k < NK; k++) \{ \\
C[i * NJ + j] += A[i * NK + k] * B[k * NJ + j]; \\
... \\
\hline
\end{tabular}
(b) CUDA kernel code
\label{tabular:2b}

\begin{tabular}{| m{8cm} |} 
\hline
\_line\_not\_to\_translate \\
\_line\_not\_to\_translate \\
\_tp0 \_op0 \_id0 ; \\
... \\
cudaMalloc ( ( \_tp0 \_op0 ) \_op1 \_id0 , sizeof ( \_tp1 ) \_op2 \_id1 \_op2 \_id2 ) ; \\
cudaMemcpy ( \_id0 , \_id1 , sizeof ( \_tp0 ) \_op0 \_id2 \_op0 \_id3 , cudaMemcpyHostToDevice ) ; \\
... \\
\hline
\end{tabular}
(c) Pre-processed CUDA host code
\label{tabular:2c}

\begin{tabular}{| m{8cm} |} 
\hline
\_\_global\_\_ \_tp0 \_id0 ( \_tp1 \_op0 \_id1 , \_tp1 \_op0 \_id2 , \_tp1 \_op0 \_id3 ) \\
\_line\_not\_to\_translate \\
\_tp0 \_id0 \_op0 blockIdx.x \_op1 blockDim.x \_op2 threadIdx.x ; \\
... \\
\_line\_not\_to\_translate \\
\_line\_not\_to\_translate \\
\_line\_not\_to\_translate \\
... \\
\hline
\end{tabular}
(d) Pre-processed CUDA kernel code
\label{tabular:2d}

\begin{tabular}{| m{8cm} |} 
\hline
\_line\_not\_to\_translate \\
\_line\_not\_to\_translate \\
cl\_mem \_id0 ; \\
... \\
\_id0 = clCreateBuffer ( context , CL\_MEM\_READ\_WRITE , sizeof ( \_tp1 ) \_op2 \_id1 \_op2 \_id2 , NULL , NULL ) ; \\
clEnqueueWriteBuffer ( command\_queue , \_id0 , CL\_TRUE , 0 , sizeof ( \_tp0 ) \_op0 \_id2 \_op0 \_id3 , \_id1 , 0 , NULL , NULL ) ; \\
... \\ \hline
\end{tabular}
(e) Host code translated by NMT system
\label{tabular:2e}

\begin{tabular}{| m{8cm} |} 
\hline
\_\_kernel \_tp0 \_id0 ( \_\_global \_tp1 \_op0 \_id1 , \_\_global \_tp1 \\
\_op0 \_id2 , \_\_global \_tp1 \_op0 \_id3 ) \\
\_line\_not\_to\_translate \\
\_tp0 \_id0 \_op0 get\_group\_id ( 0 ) \_op1 get\_group\_id ( 0 ) \\
... \\
\_line\_not\_to\_translate \\
\_line\_not\_to\_translate \\
\_line\_not\_to\_translate \\
... \\ \hline
\end{tabular}
(f) Kernel code translated by NMT system
\label{tabular:2f}
\end{figure}

\begin{figure}
\small
\centering
\begin{tabular}{| m{8cm} |} 
\hline
void mm2Cuda ( float * A , float * B , float * C ) \{ \\
cl\_mem A\_gpu ; \\
... \\
A\_gpu=clCreateBuffer(context, CL\_MEM\_READ\_WRITE, sizeof(float)*NI*NK, NULL, NULL); \\
clEnqueueWriteBuffer(command\_queue, A\_gpu, CL\_TRUE, 0, sizeof(float)*NI*NK, A, 0, NULL, NULL); \\
... \\ \hline
\end{tabular}
(g) Post-processed OpenCL host code
\label{tabular:2g}

\begin{tabular}{| m{8cm} |} 
\hline
\_\_kernel void mm2\_kernel1(\_\_global float*A, \_\_global float* B, \_\_global float* C) \{ \\
int j = get\_group\_id(0)*get\_group\_id(0)+get\_local\_id(0); \\
... \\
for ( k = 0 ; k < NK ; k ++ ) \{ \\
C[i * NJ + j] += A[i * NK + k] * B[k * NJ + j] ; \\
... \\ \hline
\end{tabular}
(h) Post-processed OpenCL kernel code
\label{tabular:2h}

\caption{Translation example of 2-D matrix multiplication code \cite{Kim}.}
\label{fig:translation_ex}
\end{figure}

\section{Limitation}
In this section, we discuss the limitation of our current approach. Fig \ref{fig:translation_mis} presents an example of sentences incorrectly translated.

\textbf{Long sentence translation} 
Despite the LSTM's ability to learn long-range temporal dependencies, we observe many sentences in which several words are truncated. In our translation, a sentence fed into the NMT system might be very long as we tokenize sentences in CUDA samples and each token becomes a word. Since many of the tokens are from function parameters, we think a new embedding layer structure needs to be developed to encode parameter parts in a different way. We leave this as future work.

\textbf{Unseen sentence translation} 
Since we manually write API usages, the dataset generated inevitably has limited function coverage and does not guarantee the successful translation of random programs. Although variable renaming and the use of an expression keyword contribute to increasing coverage with the limited number of API usages, it is still non-trivial to correctly translate a sentence not covered by API usages.

\textbf{Manual steps for API function mapping} 
We require manual efforts to write a pair of API usages. Considering that most host API functions have one-to-one correspondence between CUDA and OpenCL, most parts of the job in writing API usages would change the function name and the positions of parameters. However, since there are various sentence patterns, we still require efforts to write API usages for those sentences.

\begin{figure}
\small
\begin{tabular}{| m{2cm} || m{6cm} |} 
\hline
Source & float* A\_gpu; \_br cudaMalloc((void **) \&A\_gpu, \_expr0); \\
\hline
Translated & \_expr0 = clCreateBuffer(context, CL\_MEM\_READ\_WRITE, sizeof (DATA \_TYPE)*\_id8*\_id6, NULL, NULL); \\
\hline
Expected & cl\_mem A\_gpu; \\
\hline
\hline
Source & cudaMalloc((void ** ) \&data\_gpu, sizeof(DATA \_TYPE)*(M+1)*(N+1)); \\
\hline
Translated & data\_gpu=clCreateBuffer(context, CL \_MEM\_READ\_WRITE, sizeof(DATA \_TYPE)*\_id7*\_id6*\_id0, NULL, NULL); \\
\hline
Expected & data\_gpu=clCreateBuffer(context, CL\_MEM\_READ\_WRITE, sizeof (DATA \_TYPE)*(M+1)*(N+1), NULL, NULL); \\
\hline
\hline
Source & Convolution2D\_kernel <<< grid, block >>> (A\_gpu, B\_gpu); \\
\hline
Translated & \_clSetKernelArg("Convolution2D\_kernel" , 0, A\_gpu);
\_clSetKernelArg ("Convolution2D\_kernel", 1, B\_gpu);
\_clEnqueueNDRangeKernel(grid,  \\
\hline
Expected & \_clSetKernelArg("Convolution2D\_kernel" , 0, A\_gpu);
\_clSetKernelArg ("Convolution2D\_kernel", 1, B\_gpu);
\_clEnqueueNDRangeKernel(grid, block, "Convolution2D\_kernel"); \\
\hline
\hline
Source & cudaFree(A\_gpu); \\
\hline
Translated & clGetDevice(A\_gpu); \\
\hline
Expected & clReleaseMemObject(A\_gpu); \\
\hline
\end{tabular}
\caption{Example of sentences incorrectly translated.}
\label{fig:translation_mis}
\end{figure}
\section{Related Work}
\textbf{Using NMT for program translation}
Chen et al. \cite{treetotree} propose a novel tree-to-tree neural network and demonstrates higher accuracy for program translation, but it has a limited set of variables and restricts the vocabulary size. Zuo et al. \cite{ZuoNMT} utilize NMT techniques to deal with a cross-architecture code similarity comparison. However, it does not handle high-level language translation.

\textbf{CUDA to OpenCL translation}
Martinez et al. \cite{cu2cl} translate CUDA to OpenCL at an AST level, and Kim et al. \cite{snucu2cl} use wrapper functions to translate between CUDA and OpenCL. Compared to those works, we use source-to-source translation and exploit NMT techniques to translate programming langauges.
\section{Conclusion}
In this work, we exploited NMT techniques to translate CUDA to OpenCL. To do so, we developed a dataset generation flow and generated a dataset from CUDA benchmarks. Moreover, for training and inference phases, the pre-/post-processor were developed to enable arbitrary variable names to be translated. While our current approach correctly translates most of the sentences covered by the API usages manually written, we discovered several challenges of using NMT for program translation, especially for unseen or long sentences. In this work, the NMT itself has not been changed at all. It is our future work to improve the NMT to make it more program language translation friendly. 


\end{document}